\documentclass[10pt,twocolumn,letterpaper]{article}

\usepackage[pagenumbers]{cvpr} 

\usepackage{graphicx}
\usepackage{bm}
\usepackage{placeins}
\usepackage{float}
\usepackage{amsmath}
\usepackage{amssymb}
\usepackage{booktabs}
\usepackage{multirow}
\usepackage{algorithm}  
\usepackage{algpseudocode}
\usepackage{algorithmicx}  
\usepackage{algpseudocode} 
\usepackage[T1]{fontenc}

\usepackage[pagebackref,breaklinks,colorlinks]{hyperref}

\usepackage[capitalize]{cleveref}
\crefname{section}{Sec.}{Secs.}
\Crefname{section}{Section}{Sections}
\Crefname{table}{Table}{Tables}
\crefname{table}{Tab.}{Tabs.}

\def\cvprPaperID{403} 
\def\confName{CVPR}
\def\confYear{2022}

\begin{document}

\title{Diagnosing Batch Normalization in Class Incremental Learning}

\author{Minghao Zhou$^{1}$, Quanziang Wang$^{1}$, Jun Shu$^{1}$, Qian Zhao$^{1}$, Deyu Meng$^{1,2}$ \\ 
$^{1}$Xi'an Jiaotong University, Xi'an, China \\
$^{2}$The Macau University of Science and Technology, Macau, China\\
{\tt\small \{woshizhouminghao, sniperwqza\}@stu.xjtu.edu.cn} \quad {\tt \small xjtushujun@gmail.com}  \\ {\tt\small \{timmy.zhaoqian, dymeng\}@mail.xjtu.edu.cn}
}
\maketitle

\begin{abstract}
Extensive researches have applied deep neural networks (DNNs) in class incremental learning (Class-IL). As building blocks of DNNs, batch normalization (BN) standardizes intermediate feature maps and has been widely validated to improve training stability and convergence. However, we claim that the direct use of standard BN in Class-IL models is harmful to both the representation learning and the classifier training, thus exacerbating catastrophic forgetting. In this paper we investigate the influence of BN on Class-IL models by illustrating such BN dilemma. We further propose BN Tricks to address the issue by training a better feature extractor while eliminating classification bias. Without inviting extra hyperparameters, we apply BN Tricks to three baseline rehearsal-based methods, ER, DER++ and iCaRL. Through comprehensive experiments conducted on benchmark datasets of Seq-CIFAR-10, Seq-CIFAR-100 and Seq-Tiny-ImageNet, we show that BN Tricks can bring significant performance gains to all adopted baselines, revealing its potential generality along this line of research. 
\end{abstract}

\section{Introduction}
Deep neural networks (DNNs) have been exploited in various fields~\cite{he2016deep}. Equipped with a large training dataset, DNNs can achieve even better performance than human after extensive offline training. However, the general assumption of i.i.d. training samples does not stand in most real cases. \textit{Class incremental learning} (Class-IL)~\cite{vandeven2019three} has thus attracted much research attention, where the new classes of training data come sequentially. 
Due to the limited budget of storage of previous data, the major challenge of Class-IL is known as \textit{catastrophic forgetting}~\cite{mccloskey1989catastrophic,robins1995catastrophic}, which means that the model's knowledge of previous classes would be disrupted upon new ones' arrival. 

To mitigate catastrophic forgetting, different methods have been proposed. Regularization-based methods~\cite{aljundi2018memory,chaudhry2018riemannian,kirkpatrick2017overcoming,zenke2017continual,lee2019overcoming,li2017learning} attempt to elaborately regularize the model parameters to possibly protect the previous knowledge. Rehearsal-based methods~\cite{Chaudhry_tinyer_icml2019,buzzega2020dark,lopez2017gradient,AGEM,rebuffi2017icarl} utilize a memory buffer to include previous data information during training. Knowledge distillation\cite{rebuffi2017icarl,li2017learning,hinton2015distilling} and contrastive learning~\cite{mai2021supervised,cha2021co2l,khosla2020supervised} are also adopted to learn more discriminative and transferable representations. Another line of researches~\cite{wu2019large,belouadah2019il2m,zhao2020maintaining,hou2019learning} turn to reduce the classification bias caused by the class imbalance problem~\cite{buda2018systematic,huang2016learning} in Class-IL.

\begin{figure}[t]
  \centering
    \begin{subfigure}{0.32\linewidth}
\centering
    \includegraphics[width=0.99\linewidth]{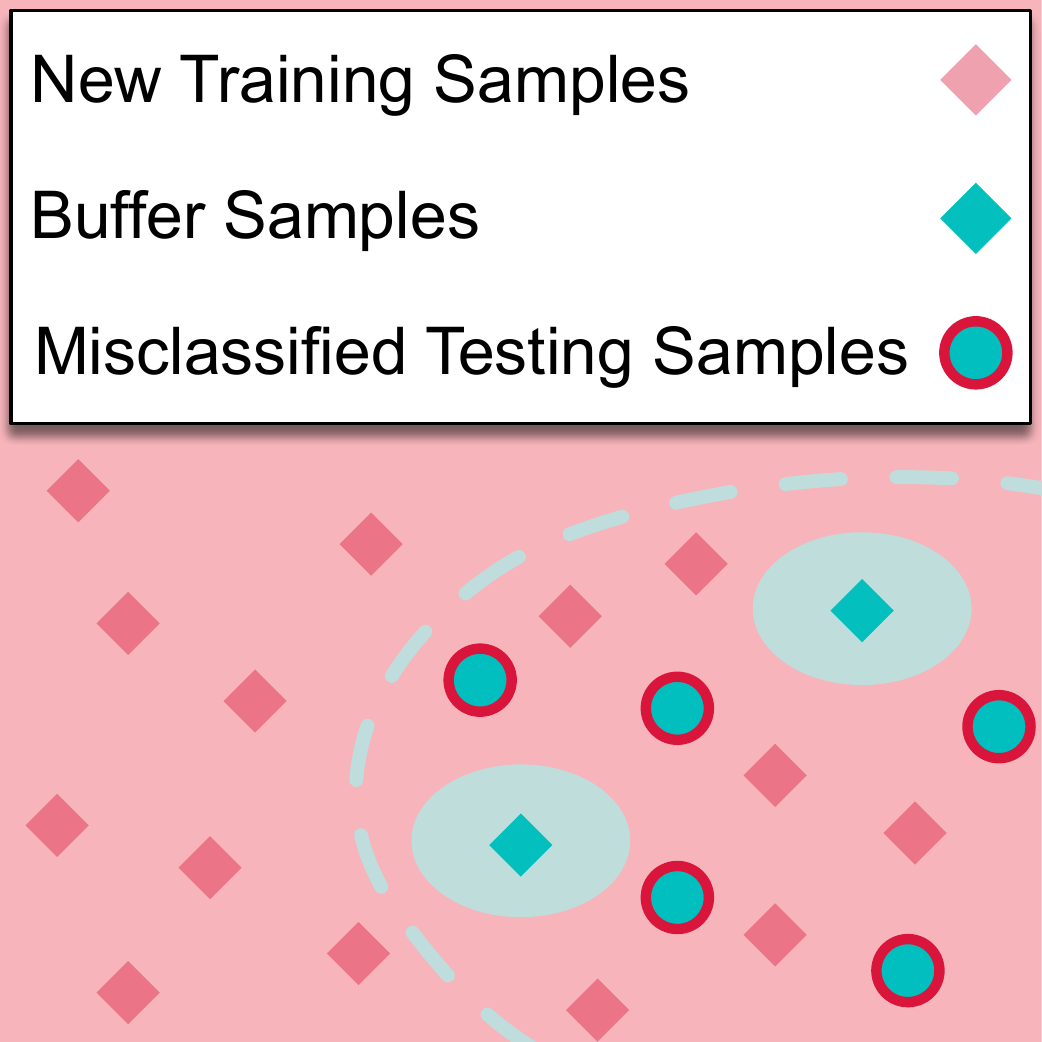}
    \caption{}
    \label{fig:1}
  \end{subfigure}
  \hfill
    \begin{subfigure}{0.32\linewidth}
\centering
    \includegraphics[width=0.99\linewidth]{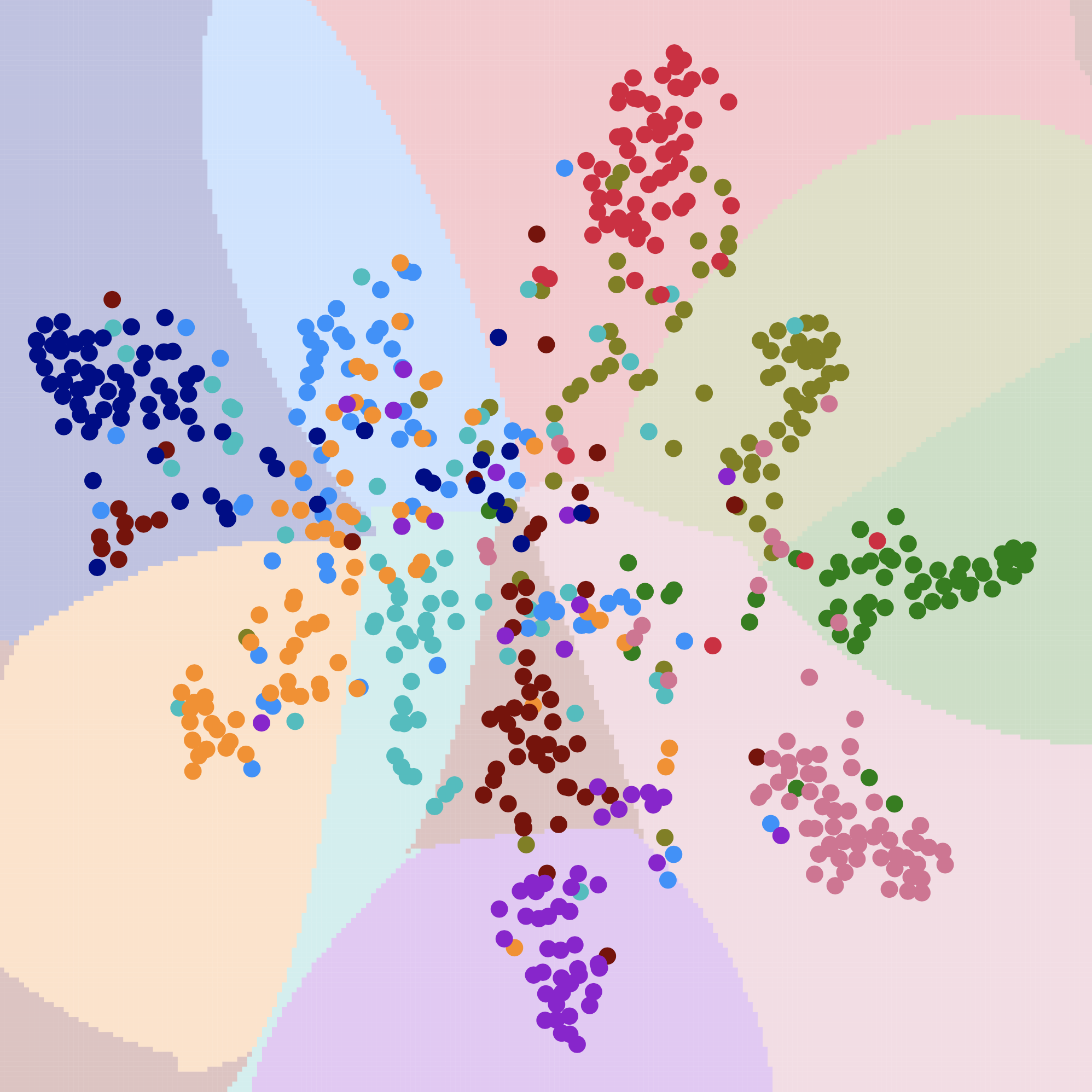}
    \caption{}
    \label{fig:balance}
  \end{subfigure}
  \hfill
    \begin{subfigure}{0.32\linewidth}
    \centering
    \includegraphics[width=0.99\linewidth]{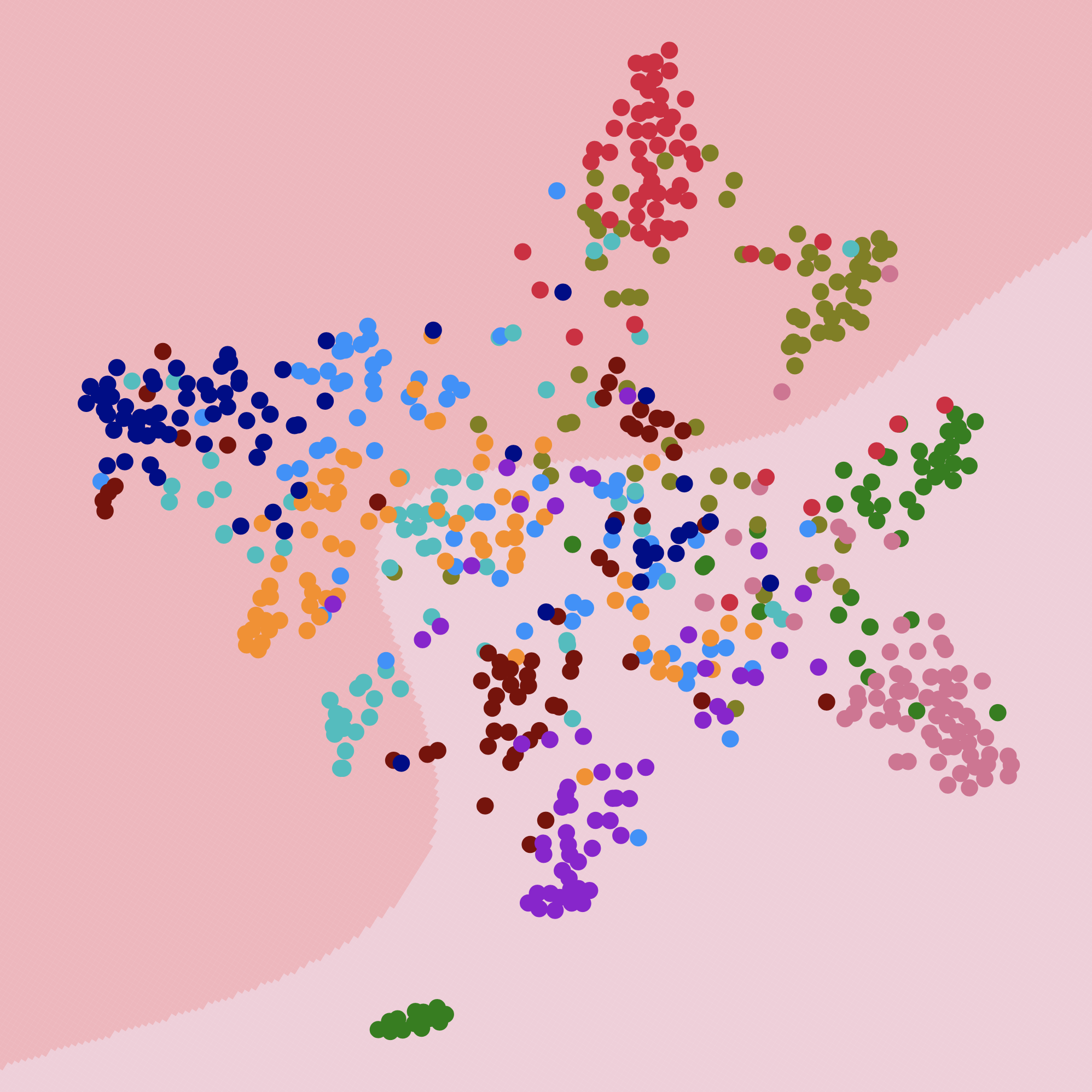}
    \caption{}
    \label{fig:current}
  \end{subfigure}
   \caption{(a) Depiction of the class imbalance issue in the overlapping area of representations, which calls for better feature learning of new classes via proper BN strategy; (b) (c) t-SNE plots of network activations of testing samples of Seq-CIFAR-10 with (b) \textbf{balanced} EMA statistics and (c) EMA statistics \textbf{biased towards the last task} (crimson and pink points). (b) and (c) share the same network parameters except for EMA statistics of BN layers. The background colors are rough prediction regions of the network\protect\footnotemark.}

   \label{fig:transfer}
  \vspace{-0.35cm}
\end{figure}
\footnotetext{The network of (b) is obtained by our proposed ER-BNT. The network of (c) is obtained by further running 100 training batches from the last task upon the network of (b), only to update the EMA statistics of BN layers. The background regions in (b) and (c) are obtained by fitting a multi-class classifier using the prediction results of the testing samples.}

While existing researches have revealed how the degraded feature extractor~\cite{mittal2021essentials} and the biased classifier~\cite{zhao2020maintaining,ahn2021ss} can lead to catastrophic forgetting, we take a different perspective to look into the issue. Specifically, we demonstrate that batch normalization (BN)~\cite{ioffe2015batch} plays a non-negligible role in causing catastrophic forgetting from the following two aspects, which are summarized as the \textit{BN dilemma}. \vspace{-0.2cm}

\begin{enumerate}
    \item  Traditional rehearsal-based methods like ER~\cite{Chaudhry_tinyer_icml2019} construct each training batch with both new training samples and buffer samples. The features of new classes tend to not be well centralized by BN layers due to the presence of massive previous task samples in the training batches, making the representations hard to be learned quickly and properly for new classes. This in turn makes their representations hardly separated from those of the previous ones, and the network is then prone to misclassify testing samples from the previous classes into the new ones owing to class imbalance issue~\cite{hou2019learning,wu2019large}. This issue is further depicted in \Cref{fig:transfer} (a), which shows that the network tends to be overfit to the buffer for previous classes in the overlapping area where the representations of buffer samples are overwhelmed by those of  abundant new training samples. 
    
    \item The deviated BN statistics easily lead to the bias of the classifier. In conventional methods, BN layers use an exponential moving average (EMA) of training batch moment statistics to replace batch moments during inference~\cite{ioffe2015batch}. We show in Figure~\ref{fig:transfer} (b) (c) that the classifier would be severely biased by using such deviated EMA statistics. This presents a new explanation for the issue that most rehearsal-free continual learning methods would lose efficacy in Class-IL. We claim that one critical insight for causing such bias is the BN discrepancy between training and inference phases~\cite{singh2019evalnorm}. 
\end{enumerate}

In this study, we firstly investigate the role of BN in Class-IL and summarize the \textit{BN dilemma} as one critical issue in obtaining better representations while maintaining fairness for Class-IL methods.
A simple solution called BN Tricks is then carefully designed for alleviating the above problems. To apply BN Tricks to ER as an example, we firstly use batches only containing new classes to train the network for better representation learning and thus protect their features from being confused with those of the previous classes. To obtain a possible unbiased classifier by eliminating the BN discrepancy, we update the network parameters and the EMA statistics separately, and use balanced batches to update the latter. As shown in Figure~\ref{fig:transfer} (b) (c), the distribution of features of different classes rarely changes with the BN statistics, which inspires us to faithfully transfer the learned representations to the inference domain of BN statistics by further training the network on buffer batches, which serve as the test-simulator. 

In summary, our main contributions are as follows: (i) We investigate the role of BN in Class-IL by illustrating the BN dilemma with extensive experiments; (ii)
 We propose an easy yet effective technique, named BN Tricks, to alleviate the BN dilemma by learning a better feature extractor while mitigating the bias of the classifier; (iii) Our method can be easily applied to general rehearsal-based methods, \eg ER~\cite{Chaudhry_tinyer_icml2019}, DER++~\cite{buzzega2020dark} and iCaRL~\cite{rebuffi2017icarl}, without inviting extra hyperparameters, so as to readily enhance their performance on various benchmarks of Class-IL.

\section{Related Work}
\subsection{Class Incremental Learning}
Various methods have been proposed to empower the models to alleviate catastrophic forgetting~\cite{delange2021continual}. Typically, regularization-based methods encourage the model to update its parameters without being far from the ``safety zone'' of previous tasks. Early approaches~\cite{kirkpatrick2017overcoming,zenke2017continual,aljundi2018memory,chaudhry2018riemannian} along this research line regularize the network parameters against changing too much according to their estimated importance weights. Inspired by knowledge distillation~\cite{hinton2015distilling}, another line of regularized methods~\cite{li2017learning,lee2019overcoming} utilize distillation loss to remind the model of previous knowledge. Regularization-based methods can work without replay exemplars. However, most of them are designed for task incremental learning (Task-IL)~\cite{vandeven2019three}, where the task indexes of the testing samples are required. They generally cannot perform well in Class-IL because the classifier would be severely biased in favour of the latest classes, without seeing exemplars of the previous classes~\cite{chaudhry2018riemannian}.

Rehearsal-based methods defy forgetting by utilizing a memory buffer composed of a small fraction of previous training samples. The most typical method is Experience Replay (ER)~\cite{Chaudhry_tinyer_icml2019}. ER updates the network with training batches consisting of samples from both new and previous classes. Such easy concatenation of the new and previous samples in ER is ill-considered because of the curse of BN as we will diagnose later. Recently, DER/DER++~\cite{buzzega2020dark} improves the performance of ER by a large margin by simply leveraging distillation loss. However, they neglect the training-inference discrepancy of BN layers, inclining to conduct degraded test performance on previous tasks. 
iCaRL~\cite{rebuffi2017icarl} focuses on representation learning and utilizes a nearest-class-mean (NCM) classifier, which has been proved to be an effective strategy for dealing with classification bias~\cite{mai2021supervised}. iCaRL mainly uses distillation loss to help memorize history knowledge. 
However, this method is also cursed by the incorrect use of BN and needs better representation learning of new classes, especially when the buffer size is large. Instead of directly using the exemplars to update the network, GEM~\cite{lopez2017gradient} and A-GEM~\cite{AGEM} keep the updating gradient in the projected direction to suppress the growing loss of the replay exemplars. These methods also tend to attain biased classifiers in Class-IL scenario. 
To reduce the classification bias, BiC~\cite{wu2019large}, IL2M~\cite{belouadah2019il2m} and WA~\cite{zhao2020maintaining} rectify the classifier in different ways. These methods yet ignore the contribution of BN to the classification bias and underestimate the contribution of feature extractors to catastrophic forgetting. 
In this study, we delve into a general amelioration for simultaneously enhancing the feature extractor and the classifier through easily rectifying the BN strategy.

\subsection{Batch Normalization}
Batch Normalization~\cite{ioffe2015batch} has been used widely in deep learning\cite{he2016deep,huang2017densely}, attributed to its capability on stabilizing the distribution of internal features and enabling faster convergence~\cite{santurkar2018does}. Variants of BN~\cite{ba2016layer,ulyanov2016instance,wu2018group} have been proposed by adopting normalization statistics of different dimensions of data other than batch statistics, and have brought performance gains in various applications. 


In recent years, the drawbacks of standard BN have attracted gradually more attention in different research areas. A significant issue of BN is its degraded performance in case of small mini-batch sizes~\cite{wu2018group,singh2019evalnorm,ioffe2017batch,du2020metanorm}. Singh and Shrivastava~\cite{singh2019evalnorm} and Ioffe~\cite{ioffe2017batch} attribute this issue to the inaccurate estimation of the EMA statistics and the training-inference discrepancy. They further propose different approaches to reduce such discrepancy for improving test performance. This line of works mainly focus on the setting where the training data distribution has no explicit shift, while we work under the setting of Class-IL, where the distribution of training data varies along sequentially coming tasks. 
In domain adaptation, BN has been modified into an adaptive manner\cite{li2018adaptive,chang2019domain} to better transfer the knowledge of the trained model to target domains. In meta learning~\cite{finn2017model}, to overcome the inconsistency between meta-train and meta-test tasks, Bronskill \etal~\cite{bronskill2020tasknorm} proposed TaskNorm as a variant of BN by combining statistics from both layer normalization~\cite{ba2016layer} and instance normalization~\cite{ulyanov2016instance} at meta-train stage with delicate design. These researches aim to transfer knowledge from learned tasks to target tasks by modifying BN information, while we deal with the challenge to behave well over all tasks using the same BN statistics during inference. In continual learning, attempts have been made to replace BN with Batch Renormalization \cite{ioffe2017batch,lomonaco2020rehearsal} and Continual Normalization \cite{pham2021continual}. Instead of replacing BN with other normalization techniques, we fix the ill application of BN in Class-IL fundamentally by changing its updating manner based on deeper understanding of catastrophic forgetting. 

\section{Preliminaries}
\subsection{Class Incremental Learning and ER}\label{sec:CLER}

In Class-IL, the training procedure is composed of online tasks with distribution shift. We assume there are totally $T$ tasks: $\{\mathcal{T}_1, \cdots, \mathcal{T}_T\}$, with each task $\mathcal{T}_t=\{(x_i^t, y_i^t)\}_{i=1}^{n_t}$, where the sample $x_i^t \in \mathcal{X}$ and the label $y_i^t \in \{ C^t_1, \cdots, C^t_{N_{t}} \}\triangleq \mathcal{C}_t$, representing the $i$-th training data pair from the $t$-th task, and $\mathcal{C}_t$ is the set of $N_t$ classes of the $t$-th task. Once finishing training on a task, its training data would be discarded, except that only a few samples from the previous tasks can be stored in a memory buffer. In Class-IL scenario, $\mathcal{C}_{t_1} \cap \mathcal{C}_{t_2} = \emptyset$ if $t_1 \neq t_2$, and the desirable mapping function is $f:\mathcal{X} \rightarrow \cup_{t=1}^T \mathcal{C}_t$. 

The goal of Class-IL is to obtain a model that behaves well on all seen tasks while forgetting less. Therefore, we evaluate all methods on two metrics: \textit{Average Accuracy} (ACC), and \textit{Backward Transfer} (BWT)~\cite{lopez2017gradient}:\vspace{-0.2cm}
\begin{equation}
\begin{split}
\text{ACC} &= \frac{1}{T}\sum_{t=1}^T R_{T, t},\\
\text{BWT} &= \frac{1}{T-1} \sum_{t=1}^{T-1}[ R_{T,t}-R_{t,t}],
\end{split}\vspace{-0.2cm}
\end{equation}
where $R_{i,j}$ is the evaluation accuracy of the model on task $\mathcal{T}_j$ after training on task $\mathcal{T}_i$. ACC can measure the overall performance of the model on all tasks, and BWT shows how much the model forgets while training on more tasks continually. Note that a very stable method with little plasticity~\cite{mermillod2013stability}, \ie, memorizes previous tasks well but learns new ones badly, can also achieve a relatively high BWT.

\begin{figure*}[t]
  \centering

  \includegraphics[width=0.9\linewidth]{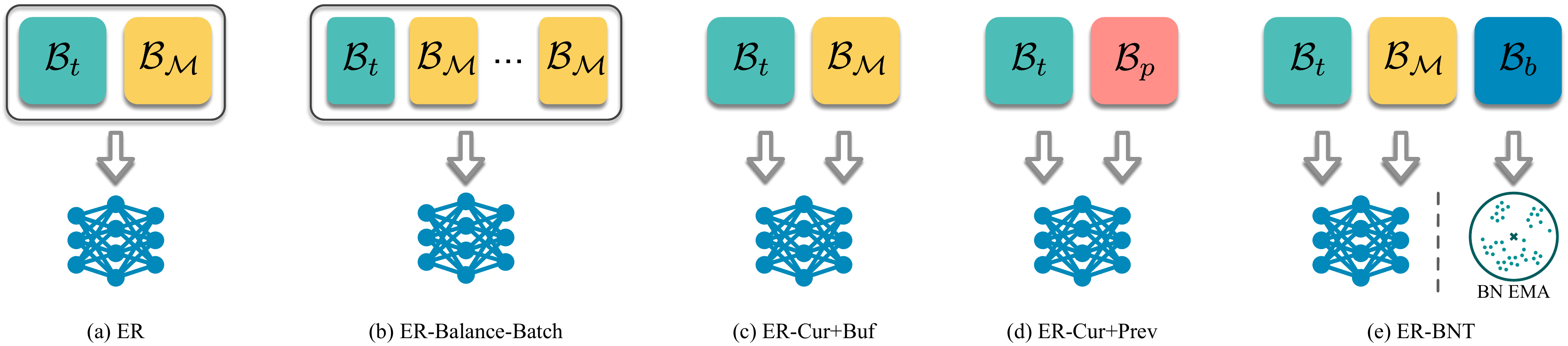}
  \vspace{-0.2cm}
\caption{Updating frameworks of (a) ER, (b)-(d) variations of ER for pilot experiments and (e) ER-BNT. In this figure, $\mathcal{B}_t, \mathcal{B_M}, \mathcal{B}_p, \mathcal{B}_b$ are current task batches, buffer batches, buffer batches only containing previous task samples and balanced batches, respectively. Except for ER-BNT, the training batches update the model parameters together with the BN EMA statistics, which is omitted in the plots.}
  \label{fig:model}
  \vspace{-0.3cm}
\end{figure*}

ER~\cite{Chaudhry_tinyer_icml2019} is a simple yet strong rehearsal-based baseline method of Class-IL. As shown in Figure~\ref{fig:model} (a), for every coming training batch $\mathcal{B}_t$ from $\mathcal{T}_t$, ER samples a batch $\mathcal{B_M}$ from the buffer, usually of the same size as $\mathcal{B}_t$. Then, ER simply concatenates $\mathcal{B}_t$ and $\mathcal{B_M}$ to feed into the network and update the model parameters, as most rehearsal-based methods~\cite{aljundi2019online,aljundi2019gradient,chaudhry2021using} do. Unless otherwise stated, we work under online Class-IL setting, where every training sample is assumed to appear only once. Thus the memory buffer should be updated upon seeing every training batch. 



\subsection{Batch Normalization} \label{sec:BN}
 In this subsection we briefly summarize the workflow of batch normalization (BN)~\cite{ioffe2015batch}. During training, the incoming feature layer of a mini-batch $\bm{x}=\{x_i\}_{i=1}^n$ is firstly standardized and then affinely transformed as follows: \vspace{-0.1cm}
 \begin{equation} \label{eqn:BN}
 \hat{\bm{x}} = \mathit{BN}(\bm{x})=\bm{\gamma} \frac{\bm{x}-\bm{\mu_{x}}}{\sqrt{\bm{\sigma}^2_{\bm{x}}+\epsilon}} + \bm{\beta}, \vspace{-0.1cm}
 \end{equation}
where $\bm{\gamma}$ and $\bm{\beta}$ are learnable parameters and can be updated by gradient descent. For application in 4-dimensional data in the format of $[B,C,H,W]$, where $B, C, H, W$ represent the batch size, channel number, width and height of the feature map, respectively, and the batch moments $\bm{\mu_x}$ and $\bm{\sigma}^2_{\bm{x}}$ are calculated by preserving only the channel dimension and taking mean/variance of $[B,H,W]$ dimensions.

 During inference, the batch-wise statistics $\bm{\mu_x}$ and $\bm{\sigma}^2_{\bm{x}}$ are no longer used, because the inference should be deterministic, \ie, the test result of each sample should be independent of other samples in the same batch. Thus, approximations of the population moments, $\bm{\mu}_r$ and $\bm{\sigma}^2_r$, are used in place of $\bm{\mu_x}$ and $\bm{\sigma}^2_{\bm{x}}$ in Eqn.~(\ref{eqn:BN}). In practice, $\bm{\mu}_r$ and $\bm{\sigma}^2_r$ are computed by recording the batch moments during training and taking exponential moving average (EMA) of them: \vspace{-0.1cm}
 \begin{equation}
 \begin{split}
\bm{\mu}_r &\gets \alpha \bm{\mu}_r + (1-\alpha) \bm{\mu_x},\\
\bm{\sigma}^2_r &\gets \alpha \bm{\sigma}^2_{\bm{r}} + (1-\alpha) \bm{\sigma}^2_{\bm{x}},
\end{split} \vspace{-0.1cm}
 \end{equation}
 where the momentum $\alpha$ is usually set as 0.9. Note that such approximation of population moments relies on the assumption that the training samples are drawn from the same distribution, which is intrinsically unsuitable for Class-IL.
 \section{Role of BN in Class-IL} \label{sec:role}
 In this section, we show the mysterious influence of BN on Class-IL models by conducting a series of pilot experiments on Seq-CIFAR-10 with buffer size equal to 500 (please refer to Section~\ref{sec:setup} for detailed experiment setup). We design three variations of ER as below, whose updating frameworks are shown in Figure~\ref{fig:model} (b)-(d).
 
  \begin{table}[t]
 \small
 \centering
 \caption{Results of the pilot experiments. } 
 \vspace{-0.2cm}
 \label{tab:pilot}
\begin{tabular}{@{}lcc@{}}
\toprule
\multicolumn{1}{c}{Method}   & ACC $(\uparrow)$ & BWT $(\uparrow)$ \\ \midrule
ER               &  61.62$\pm$1.59   &  -43.82$\pm$2.02   \\
ER-Balance-Batch &   56.40$\pm$0.80  &  -48.77$\pm$1.16   \\
ER-Cur+Buf       &  \textbf{66.83$\pm$2.00}   &   \textbf{-32.06$\pm$3.10}  \\
ER-Cur+Prev      &   62.10$\pm$2.34  &  \textcolor{red}{11.32$\pm$6.42}   \\ \bottomrule
\end{tabular}
\end{table}

\begin{figure}[t]
  \centering
\vspace{-0.2cm}
\includegraphics[width=0.99\linewidth]{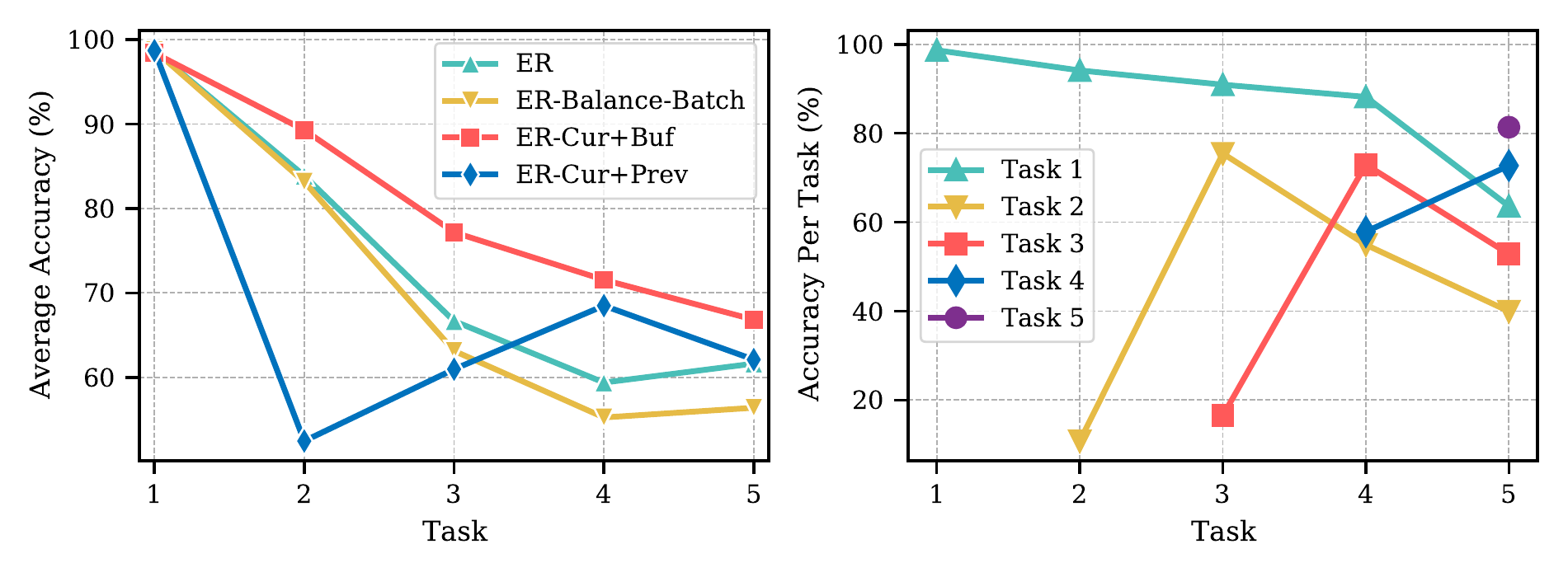} \vspace{-0.4cm}

  \caption{Left: average accuracy (ACC) of tasks during training of the pilot experiments; Right: accuracy of different tasks during training with ER-Cur+Prev. The results are averaged over 10 runs.}
  \label{fig:pilot}
  \vspace{-0.3cm}
\end{figure}

\textbf{ER-Balance-Batch} uses a balanced batch\footnote{We aim to construct training batches with the number of samples from each class approximately equal. However, our constructed batches are not exactly balanced in online memory updating setting, which is of no harm.} in each training step. To construct the balanced batch while make other conditions the same as ER, we make $t-1$ copies of $\mathcal{B_M}$ and concatenate them with $\mathcal{B}_t$, where $t>1$ is the current task index. We further offset the increased number of buffer samples by dividing their loss by $t-1$. The only difference between ER-Balance-Batch and ER is the BN statistics used during training and inference because of the different proportion of new task samples in each training batch. \textbf{ER-Cur+Buf} separately inputs $\mathcal{B}_t$ and $\mathcal{B_M}$ into the network in each training step, instead of concatenating them. We add up the losses of the two batches to update the network. Note that $\mathcal{B_M}$ also contains new task samples because of the online memory updating setting. \textbf{ER-Cur+Prev} separately inputs $\mathcal{B}_t$ and $\mathcal{B}_p$ (sampled from the buffer only containing previous task samples) into the network in each training step. In the above methods, whenever a training batch is fed into the network, the EMA statistics of the BN layers are updated conventionally. Therefore, ER-Cur+Buf and ER-Cur+Prev update the EMA statistics twice using different batches in each training step.

The results of the pilot experiments are shown in Table~\ref{tab:pilot} and Figure~\ref{fig:pilot}. ER-Balance-Batch surprisingly achieves lower ACC and worse BWT than ER. This result shows that the BN statistics indeed influence the performance of Class-IL models. The observation can be rationally explained by that in ER-Balance-Batch, the proportion of new task samples in a training batch is smaller than that in ER, and thus the batch mean is more deviated from the feature mean of new task samples and the batch variance tends to be larger. If the features of the new task samples are not properly normalized, their representations would not be well learned by the feature extractor quickly. However, how do the under-trained new classes in turn aggravate the forgetting of previous classes? This can be imputed to the class-imbalance issue in the overlapping area of representations of previous and new classes. Specifically, before training on the new task, the feature extractor could not recognize the new samples and their activations are likely to be entangled with those of previous classes. Without nudging the representations of new classes away from the previous ones via proper feature learning, the model is apt to be overfit to the buffer for previous classes, and predict testing samples whose representations are in the overlapping area into new classes. ER-Cur+Buf achieves better performance than ER and ER-Balance-Batch for this reason as well. By separately inputting $\mathcal{B}_t$ and $\mathcal{B_M}$, the current task batches $\mathcal{B}_t$ can be better normalized by their local BN statistics. This result inspires us to feed the network with batches only containing new task samples for better representation learning.

The result of ER-Cur+Prev is odd for its positive BWT and the lowest ACC at the second task (shown in Figure~\ref{fig:pilot}). By comparing ER-Cur+Prev with ER-Cur+Buf, we find that the classifier is biased without the training on $\mathcal{B_M}$. Such bias results from the \textit{BN discrepancy} between training and inference phases. Specifically, As introduced in Section~\ref{sec:BN}, BN layers use EMA statistics instead of batch moments during inference. The training BN moments of the new task samples are their batch moments, different from the EMA statistics which are computed by taking EMA of the batch moments of $\mathcal{B}_t$ and $\mathcal{B}_p$ alternately. In ER-Cur+Buf, the buffer batches $\mathcal{B_M}$, which can be seen as real-time balanced batches, stabilize the update of EMA statistics and also serve as the test-simulator to help train the classifier under balanced BN statistics. However, the BN discrepancy is not entirely eliminated in ER-Cur+Buf because the updating of $\mathcal{B}_t$ would disrupt the balance of the EMA statistics. 

In conclusion, BN affects Class-IL models from two aspects. On one hand, training on batches only containing new classes is needed for better normalization and better representation learning. On the other hand, BN discrepancy should be reduced for an unbiased classifier, which seems infeasible if imbalanced batches are used for training. Both two effects intrinsically derive the issue of the \textit{BN dilemma}. 

\begin{algorithm}[t]
\caption{ER-BNT Training Algorithm} 
\label{alg:BNT} 
\begin{algorithmic}[1] 
\Require Training data of the $t$-th task $\mathcal{T}_t$, buffer $\mathcal{M}$, model parameters $\theta$, BN EMA statistics $\phi$, learning rate $\gamma$.
\Ensure Buffer $\mathcal{M}$, model parameters $\theta$, BN EMA statistics $\phi$.
\For{$\mathcal{B}_t \sim \mathcal{T}_t$}
\State $\mathcal{B_M} \gets \mathit{sample}(\mathcal{M})$
\State $\mathcal{B}_b \gets \mathit{balance}( \mathcal{B}_t,\mathcal{B_M} )$

\State $\phi \gets \mathit{update\_EMA}(\phi, \mathcal{B}_b )$
\State without updating BN EMA statistics:
\State \quad $\mathcal{L}_t \gets \mathit{loss}_{\mathit{CE}}(\mathcal{B}_t ;\theta)$ 

\State \quad $\mathcal{L_M} \gets \mathit{loss}_{\mathit{CE}}(\mathcal{B}_\mathcal{M} ;\theta)$ 
\State $\theta \gets \theta - \gamma \ \partial (\mathcal{L}_t + \mathcal{L_M})/ \partial \theta $
\State $\mathcal{M} \gets \mathit{update\_buffer}(\mathcal{M}, \mathcal{B}_t )$
\EndFor 
\end{algorithmic}
\end{algorithm}

 \section{BN Tricks}
\subsection{ER with BN Tricks}

To alleviate the \textit{BN dilemma} described in Section~\ref{sec:role}, we propose a simple method called BN Tricks. We show the application of BN Tricks to ER, named ER-BNT, in Algorithm~\ref{alg:BNT} and Figure~\ref{fig:model} (e).

Firstly, we separate the updating of EMA statistics of BN layers from the updating of the network. Concretely, we construct a balanced batch $\mathcal{B}_b$ in every training step, by concatenating $\mathcal{B_M}$ and a fraction of $\mathcal{B}_t$ to make the number of each class approximately equal. We use $\mathcal{B}_b$ to update the EMA statistics, which is easy to implement in practice by simply forwarding $\mathcal{B}_b$ into the network without backpropagation. Secondly, to update the model parameters, we propose the joint updating framework by separately forwarding $\mathcal{B}_t$ and $\mathcal{B_M}$ into the network and add up their losses for backpropagation, meanwhile fixing the EMA statistics.  

While keeping the EMA statistics balanced, the joint updating framework is expected to finely address the \textit{BN dilemma}. $\mathcal{B}_t$ is used for better learning the representations of the new classes, with its local BN statistics. Meanwhile, to eliminate the BN discrepancy between training and inference phases, we use $\mathcal{B_M}$, which can be assumed to approximate the real-time population moments, to help fine-tune the classifier by transferring the learned representations to the inference domain of BN statistics. 

\subsection{Extension Applications} \label{sec:app}

In addition to ER, BN Tricks can also be applied to other rehearsal-based Class-IL methods. We extend BN Tricks to other two competitive methods, DER++~\cite{buzzega2020dark} and iCaRL~\cite{rebuffi2017icarl}, to show its potential generality.

\begin{algorithm}[t]
\caption{iCaRL-BNT Training Algorithm} 
\label{alg:icarlbnt} 
\begin{algorithmic}[1] 
\Require Training data of the $t$-th task $\mathcal{T}_t$, buffer $\mathcal{M}$, model parameters $\theta$, BN EMA statistics $\phi$, learning rate $\gamma$, regularization weight $\delta$.
\Ensure Buffer $\mathcal{M}$, model parameters $\theta$, BN EMA statistics $\phi$.
\State $\lambda \gets |\mathcal{M}|/|\mathcal{T}_t|$
\For{$\mathcal{B}_t \sim \mathcal{T}_t$}
\State $\mathcal{B_M}  \gets \mathit{sample}(\mathcal{M})$
\State $\mathcal{B}_b \gets \mathit{balance}( \mathcal{B}_t,\mathcal{B_M} )$
\State $\phi \gets \mathit{update\_EMA}(\phi, \mathcal{B}_b)$
\State $\mathcal{L}_b \gets \mathit{loss_{CE+KD}}(\mathcal{B}_b; \theta)$  
\State without updating BN EMA statistics:
\State \quad  $\mathcal{L}_t \gets \mathit{loss_{CE+KD}}(\mathcal{B}_t; \theta)$ 

\State $\theta \gets \theta - \gamma \ \partial (\mathcal{L}_t+\lambda \mathcal{L}_b + \delta \|\theta \|_2^2)/ \partial \theta$
\EndFor 
\State $\mathcal{M} \gets \mathit{update\_buffer}(\mathcal{M};\theta)$
\end{algorithmic}
\end{algorithm}

\begin{table*}[t]
\caption{Results of ablation study on Seq-CIFAR-10 dataset. The batches used to update EMA statistics and model parameters are shown in ``EMA'' and ``Model Parameters'' respectively. ``/'' in ``EMA'' means that we do not separately update the EMA statistics. 
}
\label{tab:ablation}
\small
\centering
\begin{tabular}{@{}lcccc@{}}
\toprule
\multicolumn{1}{c}{Methods} & EMA & Model Parameters & ACC $(\uparrow)$ & BWT $(\uparrow)$ \\ \midrule
\ ER       &     /            &    $\mathcal{B}_t \cup \mathcal{B_M}$   &  61.62$\pm$1.59   &  -43.82$\pm$2.02 \   \\
\ ER-Balance-Joint-Train    &      $\mathcal{B}_b$     &  $\mathcal{B}_t \cup \mathcal{B_M},\  \mathcal{B_M}$  &  61.44$\pm$1.29   &  -43.80$\pm$1.84 \   \\
\ ER-BNT-No-Simulator       &       $\mathcal{B}_b$    &    $\mathcal{B}_t,\  \mathcal{B}_p$    &  66.13$\pm$2.30  &   \textcolor{red}{41.76$\pm$2.04} \  \\
\ ER-BNT-Imbalance-Tracker  &     /            &    $\mathcal{B}_t,\  \mathcal{B_M}$          &   66.83$\pm$2.00  &  -32.06$\pm$3.10  \  \\
\ ER-BNT    &   $\mathcal{B}_b$     &    $\mathcal{B}_t,\  \mathcal{B_M}$      &  \textbf{72.08$\pm$0.55} &   \textbf{-23.55$\pm$0.92}  \  \\ \midrule
\ iCaRL  &   /     &  $\mathcal{B}_{t\_mix}$   &    62.33$\pm$1.17   &   -24.45$\pm$1.62   \ \\
\ iCaRL-Concat  &   /     &    $\mathcal{B}_t \cup \mathcal{B}_p$   & 62.52$\pm$0.73    &  -39.97$\pm$1.05    \ \\
\ iCaRL-BNT-No-Simulator  &   $\mathcal{B}_b$   &    $\mathcal{B}_t,\  \mathcal{B}_p$   & 59.82$\pm$1.78  &  -23.50$\pm$2.20 \ \\
\ iCaRL-BNT-Imbalance-Tracker  &   /     &    $\mathcal{B}_t,\  \mathcal{B}_b$   & 67.38$\pm$1.09  & -19.52$\pm$0.91  \ \\
\ iCaRL-BNT  &  $\mathcal{B}_b$   &    $\mathcal{B}_t,\  \mathcal{B}_b$   & \textbf{67.40$\pm$0.66}  &    \textbf{-18.46$\pm$0.55}   \ \\
\bottomrule
\end{tabular}
\vspace{-0.2cm}
\end{table*}

\textbf{DER++} adopts a different updating framework than ER. In each training step, DER++ separately feeds a current task batch $\mathcal{B}_t$ and two different buffer batches $\mathcal{B}_{\mathcal{M}_1}, \mathcal{B}_{\mathcal{M}_2}$ into the network without concatenation. The total loss is computed as \(\mathcal{L} = loss_{\mathit{CE}}(\mathcal{B}_t)+\alpha \  loss_{\mathit{KD}}(\mathcal{B}_{\mathcal{M}_1})+\beta \  loss_{\mathit{CE}}(\mathcal{B}_{\mathcal{M}_2})\), where $loss_{\mathit{CE}}$ and $loss_{\mathit{KD}}$ represent cross entropy loss and distillation loss respectively. 
Although the training framework of DER++ resembles the joint updating framework as we propose, it not only ignores the balance of EMA statistics, but also fails to clarify why the joint updating of $\mathcal{B}_t$ and $\mathcal{B_M}$ works. To apply BN Tricks to DER++, we simply modify ER-BNT by replacing $loss_{\mathit{CE}}(\mathcal{B_M})$ in ER-BNT with $\alpha \  loss_{\mathit{KD}}(\mathcal{B}_{\mathcal{M}})+\beta \  loss_{\mathit{CE}}(\mathcal{B}_{\mathcal{M}})$.




\textbf{iCaRL} adopts the Nearest-Class-Mean (NCM) classifier during inference, by which it naturally addresses the bias of the classifier made of fully-connected layers. Therefore, representation learning and preservation are critical for iCaRL, for which iCaRL utilizes distillation loss to help memorize the representations of previous classes. iCaRL works under offline Class-IL setting and it directly adds the memory buffer to the data loader of the current task before training, and samples from the constructed loader later during training. The batch moments of the sampled training batches also tend to deviate from feature moments of the new classes, making the representation learning of new classes somewhat impeded. To alleviate the problem, we present iCaRL-BNT as shown in Algorithm~\ref{alg:icarlbnt}. In each training step, we only use the constructed balanced batch $\mathcal{B}_b$ to update the EMA statistics, and separately forward $\mathcal{B}_t$ and $\mathcal{B}_b$ into the network to update its parameters. 
To offset the growing loss from buffer samples compared with original iCaRL, we multiply $\mathcal{L}_b$ by a weighting parameter $\lambda$ which is determined by the ratio of the buffer size to the size of $\mathcal{T}_t$.

\section{Experimental Results}
\subsection{Experiment Setup} \label{sec:setup}

\paragraph{Datasets.} We conduct our experiments on three datasets: Seq-CIFAR-10, Seq-CIFAR-100 and Seq-Tiny-ImageNet, for image classification. \textbf{Seq-CIFAR-10} is a sequential version of CIFAR-10 dataset~\cite{krizhevsky2009learning}. We split Seq-CIFAR-10 into 5 separate tasks and each task consists of 2 classes. Similarly, \textbf{Seq-CIFAR-100} is a sequential version of CIFAR-100 dataset~\cite{krizhevsky2009learning}. We split Seq-CIFAR-100 into 10 tasks, with each containing 10 disjoint classes. \textbf{Seq-Tiny-ImageNet} is a sequential version of Tiny-ImageNet~\cite{tinyimg} with 200 classes in total. We split Seq-Tiny-ImageNet into 10 tasks with 20 different classes in each task. We fix the task order of all the datasets in every run of training following the setup in \cite{buzzega2020dark}. \vspace{-0.3cm}

\paragraph{Implementation details.} For all the three datasets and all the implemented methods, we use an 18-layer ResNet~\cite{he2016deep} as the backbone network. We train each task for 50 epochs using SGD with learning rate equal to 0.03. We set the buffer size to 500 for Seq-CIFAR-10, 1000 for Seq-CIFAR-100 and 2000 for Seq-Tiny-ImageNet by default. To update memory buffer, we use reservoir sampling~\cite{vitter1985random} for all online Class-IL methods and use herding strategy~\cite{welling2009herding} for iCaRL(-BNT). In DER++(-BNT), we set the weighting parameters $\alpha=0.2$ and $\beta=0.5$. In iCaRL(-BNT), we set the $L_2$ regularization weight $\delta$ as $1\times 10^{-4}$. We obtain our experimental results by averaging (or computing standard deviation) over 10 runs with different random seeds.

\begin{table*}[t]
\small
\caption{Results of Class-IL benchmarks. Experiments of GEM and GSS on Seq-Tiny-ImageNet are skipped for intractable training time.} \vspace{-0.1cm}
\label{tab:cil}
\centering
\begin{tabular}{@{}lcccccc@{}}
\toprule
\multicolumn{1}{c}{\multirow{2}{*}{Methods}} & \multicolumn{2}{c}{Seq-CIFAR-10} & \multicolumn{2}{c}{Seq-CIFAR-100} & \multicolumn{2}{c}{Seq-Tiny-ImageNet} \\
& ACC $(\uparrow)$  & BWT $(\uparrow)$    & ACC $(\uparrow)$   & BWT $(\uparrow)$   & ACC $(\uparrow)$    & BWT $(\uparrow)$     \\ \midrule
Joint Training  &   91.48$\pm$0.92   &  -  &  70.51$\pm$0.72  &  -             &    58.25$\pm$0.36   &       -      \\
 SGD   &  19.64$\pm$0.04    &  -96.69$\pm$0.30  &  9.30$\pm$0.05  &-88.05$\pm$0.38   &    8.06$\pm$0.10    &   -77.52$\pm$ 0.39    \\
  LwF~\cite{li2017learning}   &   19.63$\pm$0.09   &  -96.24$\pm$0.51  &  9.60$\pm$0.16  &   -87.57$\pm$0.60    &     8.36$\pm$0.14     &  -77.17$\pm$1.46  \\
    GEM~\cite{lopez2017gradient}  & 26.86$\pm$3.59  & -82.89$\pm$4.37 & 23.50$\pm$0.57 & -71.27$\pm$0.56  &  -  &  -  \\
 A-GEM~\cite{AGEM}   &  20.43$\pm$0.42   &   -95.26$\pm$0.74   &    9.36$\pm$ 0.11   &  -87.85$\pm$0.42  &     8.10$\pm$0.12     &    -77.94$\pm$0.43    \\
  FDR~\cite{benjamin2018measuring}   &  29.10$\pm$5.02   &   -83.40$\pm$6.62   &    29.45$\pm$0.71   &  -65.33$\pm$0.81  &    19.79$\pm$0.32     &  -64.63$\pm$0.43      \\
  GSS~\cite{aljundi2019gradient}  &  48.27$\pm$2.00  & -60.25$\pm$2.55   &  14.96$\pm$0.19  &  -80.03$\pm$0.40  &  -  &  -  \\
 \midrule

 ER~\cite{Chaudhry_tinyer_icml2019} &  61.62$\pm$1.59   &   -43.82$\pm$2.02  &  27.97$\pm$0.63   &  -65.82$\pm$0.62   &   18.07$\pm$0.58  &  -65.48$\pm$0.56 \\
 ER-BNT   &  \textbf{72.08$\pm$0.55} &   \textbf{-23.55$\pm$0.92}    &  \textbf{31.61$\pm$1.07}  &   \textbf{-59.91$\pm$1.37}    &  \textbf{19.54$\pm$0.53}    &   \textbf{-64.00$\pm$0.81}   \\ \midrule
 DER++~\cite{buzzega2020dark} &  73.12$\pm$0.58   &   -23.39$\pm$1.37  &  45.89$\pm$1.03   &  -40.31$\pm$1.16   &   32.20$\pm$0.71    &  -41.95$\pm$1.53 \\
 DER++-BNT  &   \textbf{73.91$\pm$1.10}    &  \textbf{-16.50$\pm$2.57}  & \textbf{47.89$\pm$1.20} & \textbf{-29.30$\pm$1.91}      & \textbf{32.23$\pm$0.66}   &     \textbf{-29.05$\pm$1.04}      \\ \midrule
  iCaRL~\cite{rebuffi2017icarl}   &  62.33$\pm$1.17   &   -24.45$\pm$1.62   &    38.94$\pm$0.21   &    \textbf{-20.49$\pm$0.44}    &  24.72$\pm$0.44     &    \textbf{-17.82$\pm$0.42}       \\
    iCaRL-BNT  &    \textbf{67.40$\pm$0.66}  &    \textbf{-18.46$\pm$0.55}   &   \textbf{42.18$\pm$0.47}  & -21.41$\pm$0.53    &  \textbf{26.50$\pm$0.28}  & -18.16$\pm$0.39    \\
 \bottomrule
\end{tabular}
\vspace{-0.3cm}
\end{table*}

\begin{figure*}
  \centering
    \begin{subfigure}{0.33\linewidth}
\centering
      \includegraphics[width=0.78\linewidth]{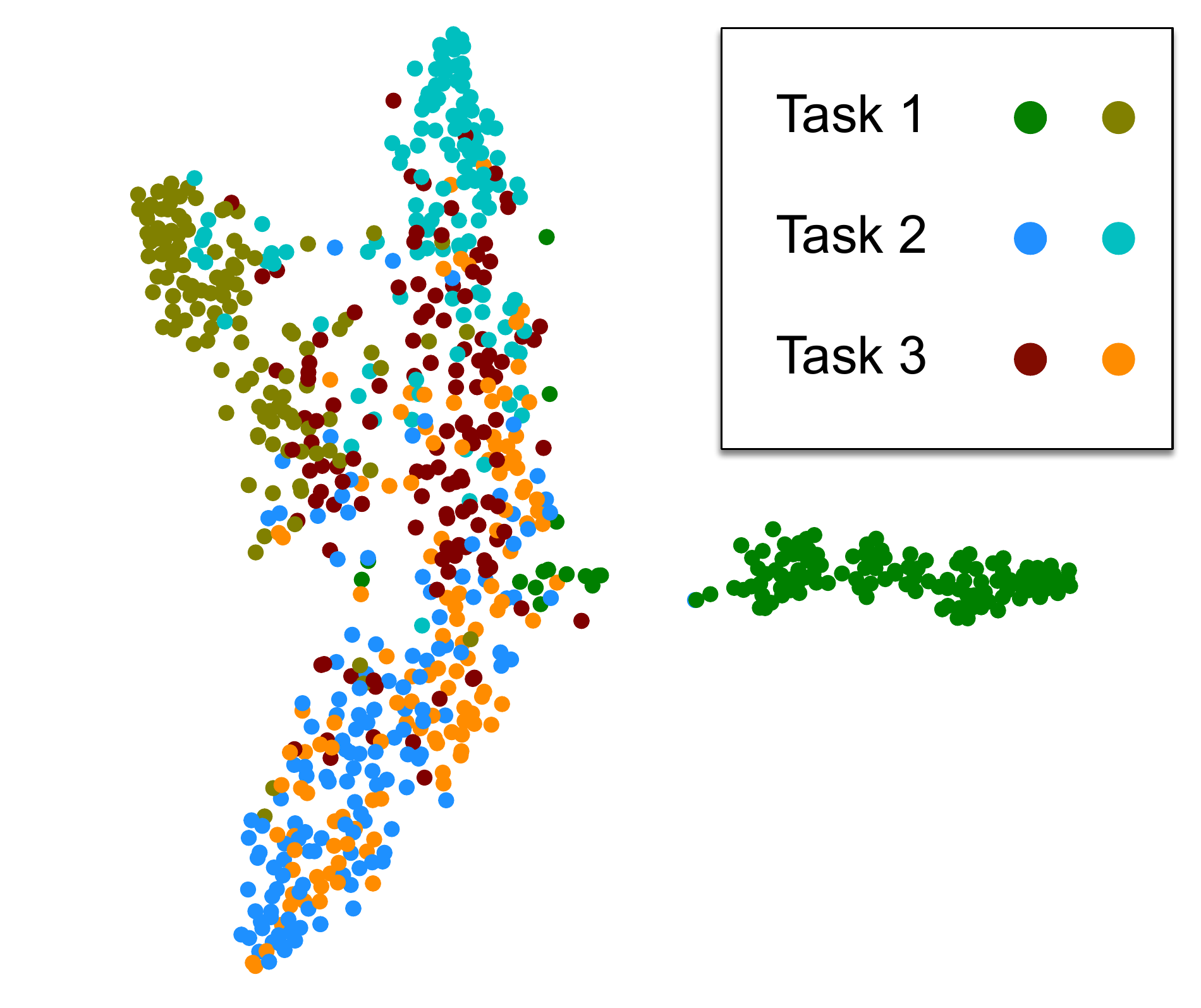}
    \caption{}
    \label{fig:mix_tsne}
  \end{subfigure}
  \hfill
    \begin{subfigure}{0.33\linewidth}
\centering
      \includegraphics[width=0.78\linewidth]{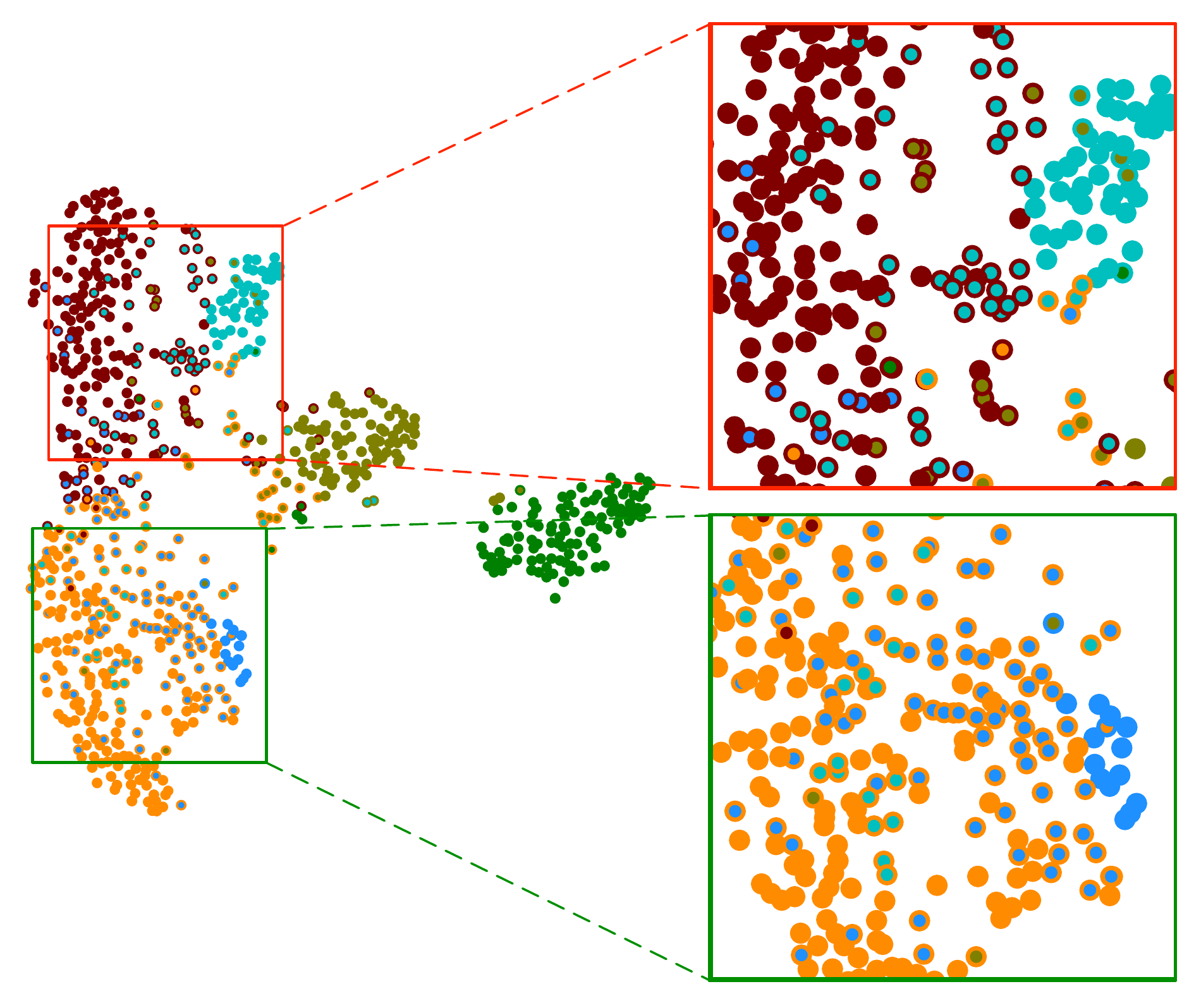}
    \caption{}
    \label{fig:er_tsne}
  \end{subfigure}
  \hfill
    \begin{subfigure}{0.33\linewidth}
    \centering
    \includegraphics[width=0.78\linewidth]{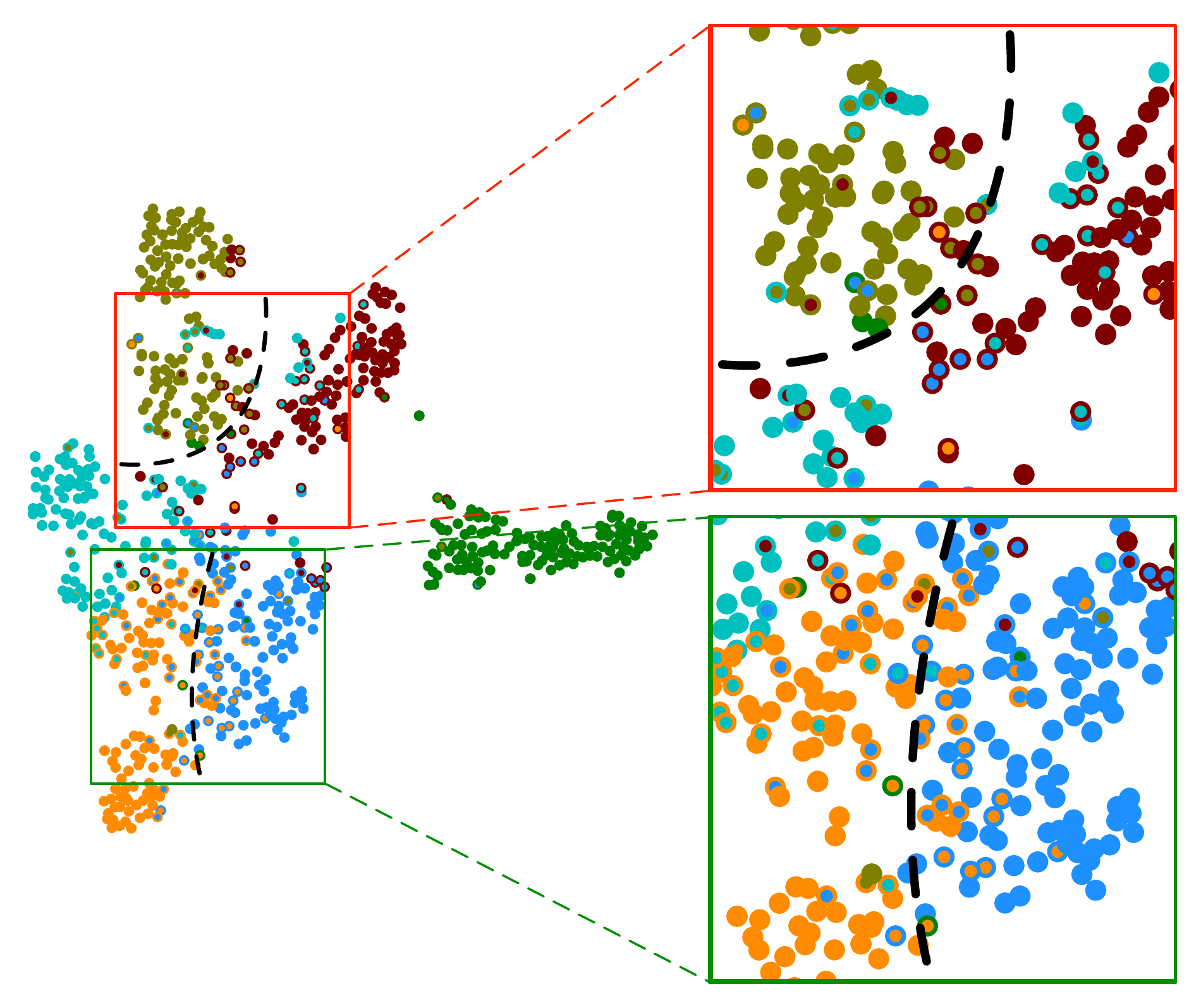}
    \caption{}
    \label{fig:er_bnt_tsne}
  \end{subfigure}
\vspace{-0.2cm}
  \caption{(a) t-SNE plot of network activations of samples from the first 3 tasks of Seq-CIFAR-10 \textbf{before} training on the third task; (b) (c) t-SNE plots \textbf{after} training the third task using \textbf{ER} and \textbf{ER-BNT} respectively. The face color of each point indicates the label of the sample and the edge color indicates the prediction by the trained network. Best viewed in color. }
  \label{fig:tsne}
  \vspace{-0.4cm}
\end{figure*}

\begin{figure}[t]
  \centering
  \includegraphics[width=0.99\linewidth]{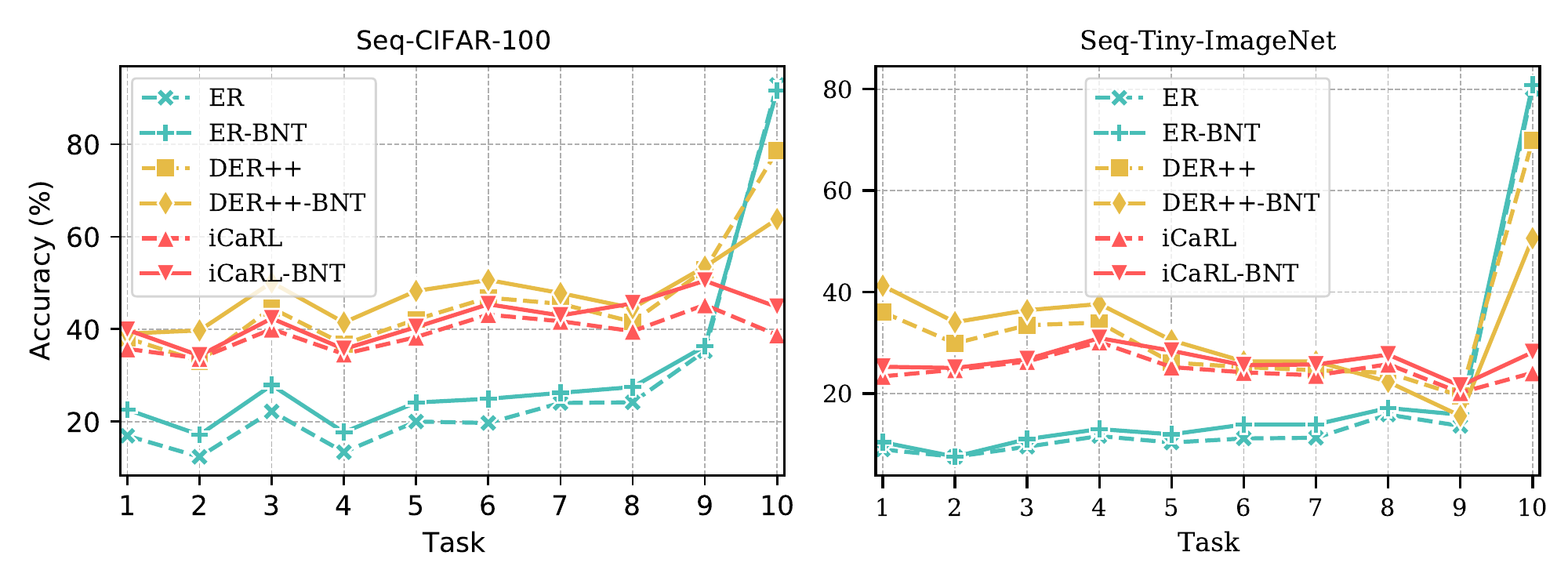}
  \vspace{-0.4cm}
  \caption{Final test accuracy on 10 tasks of Seq-CIFAR-100 and Seq-Tiny-ImageNet respectively, averaged over 10 runs.}
  \label{fig:zhexian}
  \vspace{-0.4cm}
\end{figure}

\subsection{Effect of BN Dilemma} \label{subsec:effect}

In this subsection, we demonstrate the effect of \textit{BN dilemma} with a series of ablation experiments on Seq-CIFAR-10. The results are quantitatively reported in Table~\ref{tab:ablation} and visually demonstrated in Figure~\ref{fig:tsne}. \vspace{-0.3cm}
\paragraph{Ablation study of ER-BNT.}
\textbf{ER-Balance-Joint-Train} is designed in order to show the transfer of knowledge between different inference BN statistics. This strategy mainly uses the concatenated batches to update the feature extractor, as ER does, and it transfers the learned knowledge to the balanced inference BN statistics by training with $\mathcal{B_M}$. 
The performance of ER-Balance-Joint-Train is similar to that of ER, which shows the feasibility of the transfer. The comparison between ER-Balance-Joint-Train and ER-BNT shows that training with $\mathcal{B}_t$ without concatenation with $\mathcal{B_M}$ can indeed help learning better representations of the new classes. \textbf{ER-BNT-No-Simulator} is implemented by removing the new task samples from $\mathcal{B_M}$ in the training of ER-BNT. Similar to ER-Cur+Prev in Section~\ref{sec:role}, the accuracy on latest tasks is extremely low, which results in its positive BWT. This experiment clearly shows the critical influence of the BN discrepancy. Therefore, the training on $\mathcal{B_M}$ which serves as the test-simulator is indispensable. \textbf{ER-BNT-Imbalance-Tracker} is the same as ER-Cur+Buf implemented in Section~\ref{sec:role}. The result shows that $\mathcal{B}_t$ would disrupt the balanced EMA statistics, substantiating the necessity of the separated updating framework.\vspace{-0.3cm}

\begin{table*}[t]
\small
\centering
\caption{Experiments on Seq-CIFAR-100 with different buffer sizes.\vspace{-0.1cm}}
\label{tab:buf}
\begin{tabular}{@{}lcccccc@{}} 
\toprule
\multicolumn{1}{c}{\multirow{2}{*}{Methods}} & \multicolumn{2}{c}{$|\mathcal{M}|=200$} & \multicolumn{2}{c}{$|\mathcal{M}|=500$} & \multicolumn{2}{c}{$|\mathcal{M}|=5000$} \\
& ACC $(\uparrow)$  & BWT $(\uparrow)$    & ACC $(\uparrow)$   & BWT $(\uparrow)$   & ACC $(\uparrow)$    & BWT $(\uparrow)$     \\ \midrule
 ER &  13.94$\pm$0.58 &   -81.54$\pm$0.60  &   20.34$\pm$0.51  &  -74.38$\pm$0.69   &   49.35$\pm$0.61  &  -39.93$\pm$0.64\\
 ER-BNT   &  \textbf{14.60$\pm$0.50}  &  \textbf{-81.07$\pm$0.63}   & \textbf{22.26$\pm$0.89}  & \textbf{-71.66$\pm$1.19}   & \textbf{55.21$\pm$0.58}  &  \textbf{-26.74$\pm$0.83}   \\ \midrule
 DER++  &  24.39$\pm$1.85  & -67.72$\pm$3.25 & 36.51$\pm$1.65   &  -51.54$\pm$2.40  &    59.49$\pm$0.68    &  -23.20$\pm$1.55   \\ 
 DER++-BNT  &  \textbf{28.60$\pm$1.38}  &   \textbf{-54.37$\pm$3.59 }   &  \textbf{40.43$\pm$0.83}  &   \textbf{-38.15$\pm$2.12}   & \textbf{59.84$\pm$0.62}  &  \textbf{-16.07$\pm$0.89}   \\ \midrule
  iCaRL   &  32.92$\pm$0.68  &    \textbf{-26.38$\pm$0.88}   &  37.48$\pm$0.46  & \textbf{-22.21$\pm$0.72}   &  40.81$\pm$0.49 &  -17.86$\pm$0.52   \\ 
  iCaRL-BNT  &  \textbf{33.29$\pm$0.71} &  -28.17$\pm$0.71  &  \textbf{39.17$\pm$0.37}  &   -23.41$\pm$0.47   &  \textbf{48.78$\pm$0.32} &  \textbf{-15.81$\pm$0.29}  \\ 
 \bottomrule
\end{tabular}
\vspace{-0.3cm}
\end{table*}

\paragraph{Ablation study of iCaRL-BNT.}
As mentioned in Section~\ref{sec:app}, iCaRL adds the memory buffer to the training data loader before training and samples from the newly constructed data loader during training. We denote the sampled batches as $\mathcal{B}_{t\_mix}$, where the proportion of previous samples in each batch is determined by the buffer size and the number of training data. Usually, the number of previous samples in each training batch is much smaller than that of new samples. We implement \textbf{iCaRL-Concat} following the updating framework of ER. iCaRL-Concat achieves similar ACC with iCaRL but significantly lower BWT. We can infer from the result of iCaRL-Concat that the representation learning of new classes are hampered because of the concatenation of current and buffer batches, which would eventually lead to more serious forgetting. Note that the distillation loss in iCaRL allows it to revisit exemplars less frequently while preserve their representations. Similar to the ablation study of ER-BNT, we implement \textbf{iCaRL-BNT-No-Simulator} and \textbf{iCaRL-BNT-Imbalance-Tracker} to emphasize the importance of the test-simulator and the separated updating of model parameters and EMA statistics. \vspace{-0.4cm}

\paragraph{Visualization of BN dilemma.}
Figure~\ref{fig:tsne} (a) is the t-SNE plot of network activations of testing samples from the first 3 tasks before training on the third task. It can be seen that the activations of samples from the new task overlap those from the second task. After training on the third task with ER (see Figure~\ref{fig:tsne} (b)), the representations of samples from the second and the third tasks are still entangled with each other. What's worse, the network misclassifies most of the previous samples in the overlapping area into new classes. We visualize the network activations of ER-BNT in Figure~\ref{fig:tsne} (c). Although the representations of new and previous classes overlap to some mild extent, the classifier would not misclassify most previous samples in the overlapping area into new classes, thus exhibiting clear classification boundaries. The visualization vividly shows that ER-BNT possesses better discrimination and fairness.

\begin{figure}[t]
  \centering
  \includegraphics[width=0.99\linewidth]{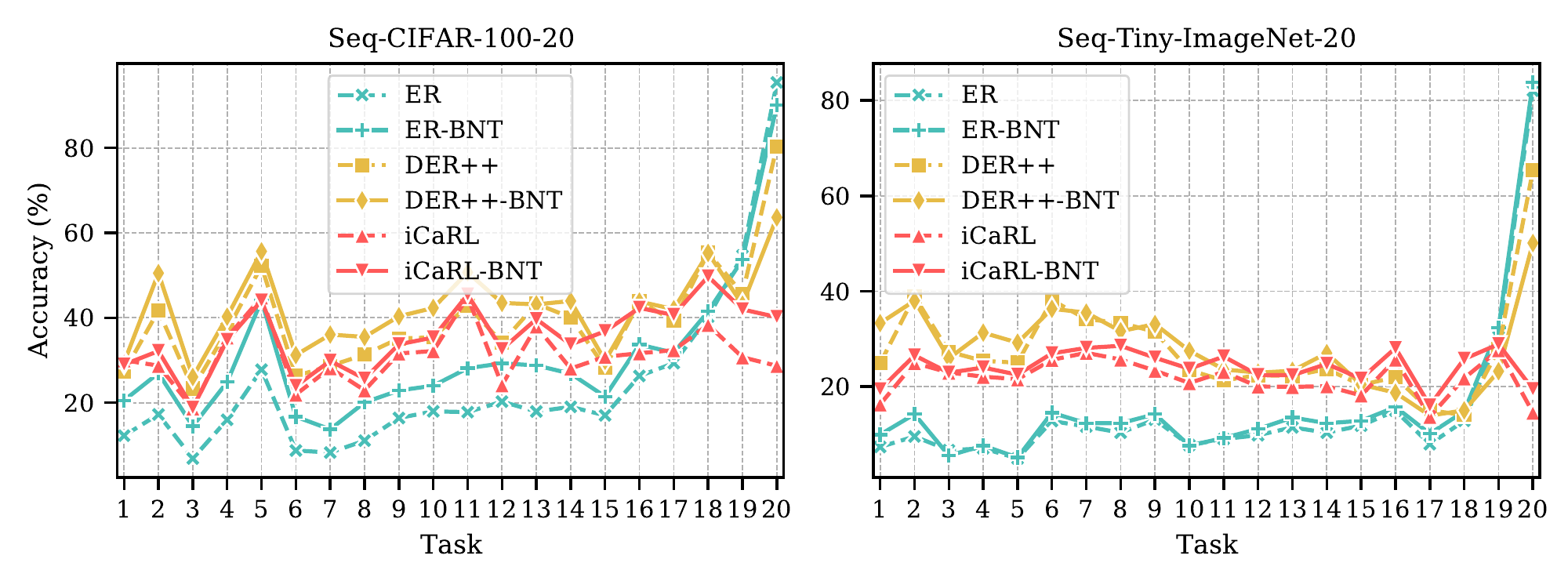}
  \vspace{-0.3cm}
  \caption{Final test accuracy on 20 tasks of Seq-CIFAR-100-20 and Seq-Tiny-ImageNet-20 respectively, averaged over 10 runs.}
  \label{fig:zhexian2}
  \vspace{-0.4cm}
\end{figure}

\subsection{Evaluation of BN Tricks}
To evaluate our proposed BN Tricks, we firstly perform experiments of its applications to ER, DER++ and iCaRL on three benchmark datasets. We also compare our methods with other Class-IL methods, including LwF~\cite{li2017learning}, GEM~\cite{lopez2017gradient}, A-GEM~\cite{AGEM}, FDR~\cite{benjamin2018measuring} and GSS~\cite{aljundi2019gradient}. We also provide the upper bound (Joint Training) by training all tasks jointly, and the lower bound (SGD) by simply training the network on the data stream. 

The results of ACC and BWT of all the implemented methods are reported in Table~\ref{tab:cil}. 
For LwF, GEM and A-GEM, the forgetting is serious for their low accuracy on previous tasks. We claim that the drift of EMA statistics is the main cause for their classification bias, similar to the effect shown in Figure~\ref{fig:transfer} (c). For rehearsal-based methods FDR and GSS, BN Tricks can also be applied to them but we have not included their results since they are not competitive enough as our adopted baseline methods. ER-BNT outperforms ER in all benchmarks with respect to ACC and BWT. In Seq-CIFAR-10 where the total number of classes is smaller, ER-BNT achieves 17\% gain in ACC and reduces BWT by 46\%. DER++-BNT also improves DER++ in BWT significantly. As shown in Figure~\ref{fig:zhexian}, DER++-BNT achieves higher accuracy of previous tasks, at the sacrifice of the decreased accuracy of the last task, which is acceptable. For iCaRL-BNT, although its BWT for Seq-CIFAR-100 and Seq-Tiny-ImageNet is lower than iCaRL, we claim the reason is that the metric BWT would reward models that are stable but implastic. As shown in Figure~\ref{fig:zhexian}, the accuracy on every task of iCaRL is lower than that of iCaRL-BNT at final evaluation, which is also the case when finishing training the previous tasks. The low accuracy of iCaRL on latest tasks makes it seem to forget less when calculating BWT. 

We further study the impact of buffer sizes on the performance of BN Tricks. We conduct experiments of BN Tricks applied to three baseline methods on Seq-CIFAR-100 with different buffer sizes, as reported in Table~\ref{tab:buf}. Please also refer to the results in Table~\ref{tab:cil}, where we use the buffer size of 1000 for Seq-CIFAR-100. The results show that the baseline methods with BN Tricks can achieve consistent performance gains with small or large storage budget. iCaRL-BNT improves iCaRL more when the buffer size is larger, because in this case previous samples appear more frequently in iCaRL's training batches.

To evaluate BN Tricks with longer task sequences, we construct Seq-CIFAR-100-20 and Seq-Tiny-ImageNet-20 by splitting CIFAR-100 and Tiny-ImageNet into 20 tasks, with each task containing 5 and 10 classes, respectively. We remain the other experiment setup and show the results of BN Tricks evaluated on the two datasets in Figure~\ref{fig:zhexian2}, where the superiority of BN Tricks can also be clearly observed. 

To sum up, by evaluating the three applications of BN Tricks and comparing them with other methods on different benchmarks, the importance of BN in Class-IL models and the effectiveness of BN Tricks can be substantiated. More experimental results are provided in the supplementary material for more comprehensive verification.

\section{Conclusion}
In this work, we have revisited the critical issue of better discrimination and fairness in Class-IL from the perspective of batch normalization. We illustrate the BN dilemma and thereafter address this issue with our proposed BN Tricks, which is fully compatible with general rehearsal-based Class-IL methods. We further validate our understandings with extensive experiments. This study is potentially useful to inspire future researches on Class-IL, or other scenarios of continual learning, to be aware of the use of BN and consider to incorporate BN Tricks in the implementations of more general baseline methods.

{\small
\bibliographystyle{ieee_fullname}
\bibliography{egbib}
}

\end{document}